\documentclass[sigconf]{acmart}
\usepackage{hhline}
\usepackage{multirow}
\usepackage{bbding}
\usepackage{adjustbox}
\usepackage{subfig}
\usepackage[title]{appendix}
\AtBeginDocument{%
  \providecommand\BibTeX{{%
    \normalfont B\kern-0.5em{\scshape i\kern-0.25em b}\kern-0.8em\TeX}}}

\setcopyright{acmcopyright}
\copyrightyear{2023}
\acmYear{2023}
\acmDOI{XXXXXXX.XXXXXXX}

\acmPrice{15.00}
\acmISBN{978-1-4503-XXXX-X/18/06}

\renewcommand\footnotetextcopyrightpermission[1]{}
\begin{document}
\fancyhead{}

\title{Vision-Language Pre-training with Object Contrastive Learning for 3D Scene Understanding}

\author{Taolin Zhang$^{1*}$, Sunan He$^{1*}$, Tao Dai$^{2\boxtimes}$, Bin Chen$^{3}$, Zhi Wang$^{2}$ and Shu-Tao Xia$^{1,4}$}
\thanks{$^\ast$Equal contribution.}
% \thanks{$^\dagger$Work done during an internship at Tencent.}
\thanks{$^\boxtimes$Corresponding author: Tao Dai.}
\affiliation{%
	\institution{$^1$Tsinghua Shenzhen International Graduate School, Tsinghua University}
	\institution{$^2$College of Computer Science and Software Engineering, Shenzhen University}
    \institution{$^3$Harbin Institute of Technology, Shenzhen}
    \institution{$^4$Research Center of Artificial Intelligence, Peng Cheng Laboratory}
    \country{}
}
\email{{zhangtlin3, daitao.edu}@gmail.com, {hsn20,wangzhi}@mails.tsinghua.edu.cn, chenbin2021@hit.edu.cn, xiast@sz.tsinghua.edu.cn}

\begin{abstract}
    In recent years, vision language pre-training frameworks have made significant progress in natural language processing and computer vision, achieving remarkable performance improvement on various downstream tasks. However, when extended to point cloud data, existing works mainly focus on building task-specific models, and fail to extract universal 3D vision-language embedding that generalize well. We carefully investigate three common tasks in semantic 3D scene understanding, and derive key insights into the development of a pre-training model.
    Motivated by these observations, we propose a vision-language pre-training framework 3DVLP (3D vision-language pre-training with object contrastive learning),  which transfers flexibly on 3D vision-language downstream tasks.  3DVLP takes visual grounding as the proxy task and introduces Object-level IoU-guided Detection (OID) loss to obtain high-quality proposals in the scene. Moreover, we design Object-level Cross-Contrastive alignment (OCC) task and Object-level Self-Contrastive learning (OSC) task to align the objects with descriptions and distinguish different objects in the scene, respectively.
    Extensive experiments verify the excellent performance of 3DVLP on three 3D vision-language tasks, reflecting its superiority in semantic 3D scene understanding.
\end{abstract}





\maketitle

\begin{figure}[h]
    \centering
    \includegraphics[width=0.5\textwidth]{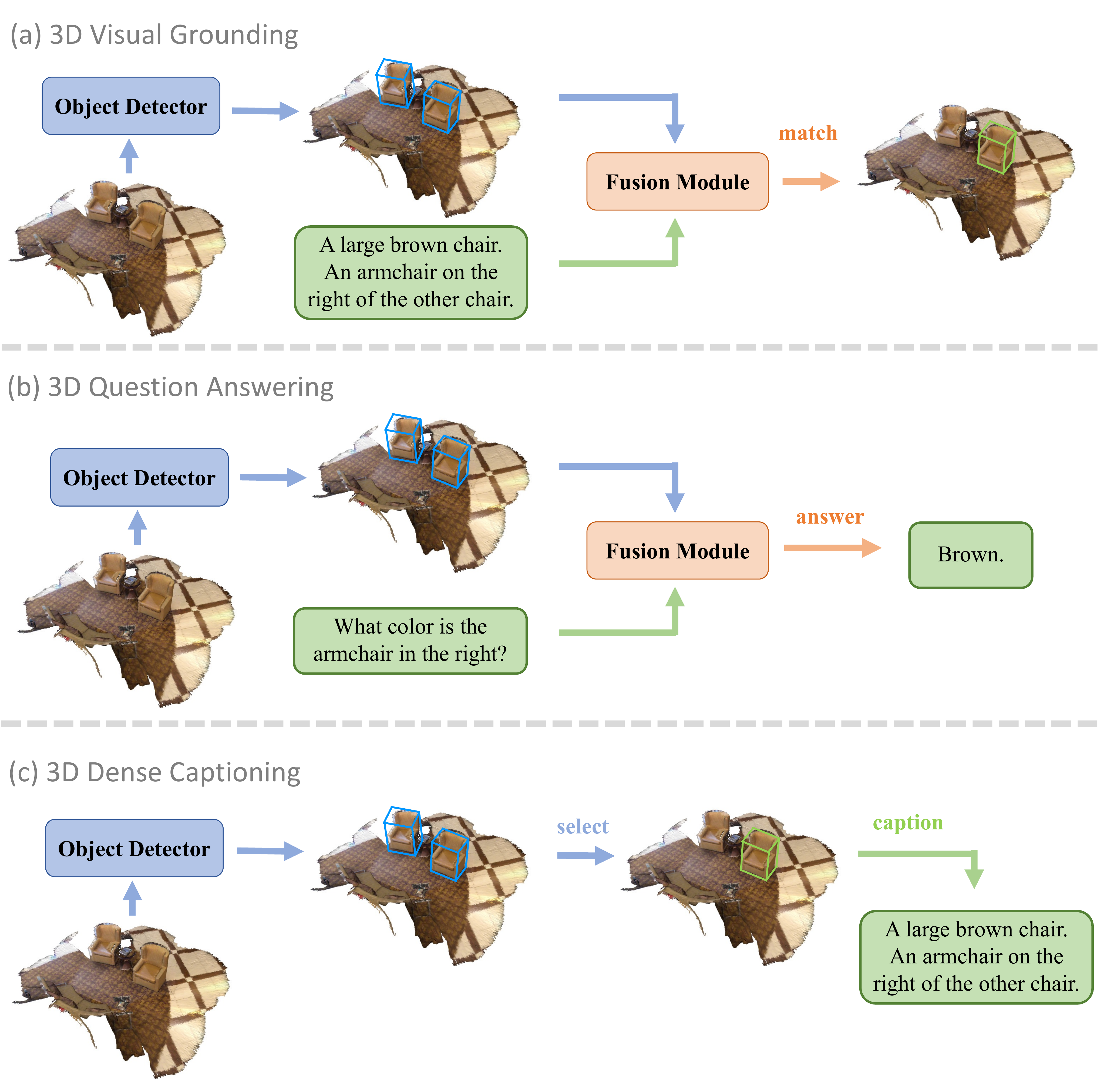}
    \caption{Relationship between 3D vision-language tasks. Firstly, all the tasks rely heavily on the object detector to locate object in the scene.  Secondly, 3D vision-language tasks require an effective fusion module to understand the connection between point cloud and language.}
    \label{task}
\end{figure}

\section{Introduction}
Semantic 3D scene understanding has recently attracted increasing research interest due to its
wide applications such as automatic driving, human-machine interaction, \textit{etc}.
Much progress has been made in semantic 3D scene understanding, with task-specific models continuously pushing the state-of-the-art in various downstream tasks including visual grounding \cite{chen2020scanrefer,zhao20213Dvg,cai20223Djcg}, dense captioning \cite{chen2021scan2cap}, and question answering \cite{azuma2022scanqa}.

While effective on their respective benchmarks, the task-specific representations obtained by existing approaches prevent them from generalizing well to other tasks. A common practice for extracting joint multimodal representation is to adopt the pre-training plus fine-tuning paradigm, whose effectiveness have been demonstrated by the remarkable success in 2D vision-language pre-training \cite{li2021align,chen2020uniter,li2019visualbert,tan2019lxmert,zhai2022lit,alayrac2022flamingo}. Existing works on 3D vision-language pre-training are still limited, which motivates us to introduce this paradigm into semantic 3D scene understanding in an appropriate way. However, 3D vision-language pre-training differs from pre-training in NLP and 2D vision-language tasks since point cloud data is introduced \cite{guo2020deep}. The task-agnostic objectives designed in previous works cannot be directly applied to 3D vision-language pre-training due to the gap of downstream tasks.
In light of these consideration, it is essential to identify the shared nature across different tasks in semantic 3D scene understanding to further determine the appropriate pre-training model.

% Although pre-training have been successful in computer vision and natural language processing, . Existing works primarily focus on end-to-end learning, resulting in highly task-specific models that fail to build the general relationship across modalities in 3D vision-language multimodal learning.
% However, it has been observed that different tasks have different targets though, they are highly correlated and can be united based on a pre-trained model.

% Common tasks in semantic 3D scene understanding, which involve effective fusion of information from point clouds and language, currently include visual grounding \cite{chen2020scanrefer}, dense captioning \cite{chen2021scan2cap}, and question answering \cite{azuma2022scanqa}. 

Figure \ref{task} provides an intuitive depiction of the relationships among three 3D vision-language tasks.
% It can be seen that these tasks differ from traditional 2D pre-training model downstream tasks as they bring new challenges in modeling the scene. Point cloud data contains information about multiple objects in the scene, and the language description data also provides information about the relationships between objects in the scene.
% The 3D visual grounding task takes point cloud scene data and a descriptive expression as input and outputs the bounding box of the target object. In contrast, the 3D dense captioning task takes the scene data and captions for all objects. The 3D question answering task uses the scene data to generate an answer to a question. 
Two key observations emerages from the comparision of these tasks. Firstly, all of these tasks rely heavily on the object detection when applying two-stage pipeline models, which is a common practice in semantic 3D scene understanding
\cite{chen2020scanrefer,chen2021scan2cap}. Secondly, 3D vision-language tasks require an effective fusion module to enable information interaction between point cloud and language  for a deeper understanding of the relationships between objects in the scene, such as the matching stage in the visual grounding \cite{zhao20213Dvg,cai20223Djcg} and the classification of answers in the question answering \cite{azuma2022scanqa}.

These observations in semantic 3D scene understanding pose several challenges in designing an effective training paradigm for the pre-training model to obtain universal embeddings and achieve better transfer performance flexibly in downstream tasks.
Firstly, high-quality bounding boxes are required for object detection, which can be further fed into task-specific heads in downstream tasks. These boxes represent the model's ability to segment the scene at the object level, as demonstrated by works that use a detection-then-matching pipeline \cite{cai20223Djcg,chen2020scanrefer,zhao20213Dvg,achlioptas2020referit3D}.
Secondly, object detection requires the model to distinguish between different objects in the scene, especially when there are many objects similar to the target, which is common in real-life situations \cite{chen2020scanrefer}. This means the model needs to be able to identify what makes objects distinct in the scene, which is a challenging task that has not yet been fully addressed.
Thirdly, the fusion module suffers from the issue that the data come from different modalities are unaligned, as similar to the cross-modal problems in 2D vision language learning \cite{li2021align,chen2020uniter}. Point cloud features and word token embeddings exist in different spaces, making it challenging for the fusion module to model their interactions.

To this end, we propose 3DVLP: vision-language pre-training with object contrastive learning in semantic 3D scene understanding. 3DVLP is the first pre-training framework that effectively addresses the challenges mentioned above.
(1) To obtain better object bounding boxes, we introduce \textbf{Object-level IoU-guided Detection} (OID) loss in our pre-training pipeline. Specifically, we leverage visual grounding as the proxy task, as it shares the same objective of localizing high-quality bounding boxes. Additionally, we incorporate Distance IoU (DIoU) loss \cite{zheng2020distance} and label smoothing in the matching stage at the object level to achieve faster convergence and better performance.
% by a distance-aware IoU metric. Additionally, we observe that previous work considers the matching stage as a classification task, and only the predicted box with the highest IoU with the ground truth is optimized\cite{zhao20213Dvg,cai20223Djcg}, limiting the effectiveness of the DIoU loss since there may be multiple bounding boxes targeting the same object. Therefore, we 
% and optimize the model using all detection boxes with IoU greater than a specific threshold in early training, 
(2) We further introduce \textbf{Object-level Self-Contrastive learning} (OSC) task to distinguish the target object from others.
The self-contrastive learning is performed at the object level, where boxes with an IoU higher than a specific threshold are considered positive samples, while others are regarded as negative ones.
This self-contrastive loss is designed to bring positive samples closer to each other and far away from the negative ones.
(3) To enable fully information intereaction between point cloud and language, we further design \textbf{Object-level Cross-Contrastive alignment} (OCC) task as a proxy task to align the unimodal representation across these two modalities.
We use a similar IoU filter as in OSC to generate positive and negative samples, which are then fed as inputs to calculate the cross-contrastive loss.
The cross-contrastive loss is introduced to pull the embedding of positive samples closer to the anchor feature of the target language description.

% This article aims to explore issues related to 3D vision-language multimodal pre-training, with a focus on suitable supervised pre-training objectives and task transfering. Specifically, we will investigate how supervised pre-training models based on visual grounding can be utilized to process point cloud and language  data, and how models can be fine-tuned based on the task-agnostic backbone to achieve excellent performance in various downstream tasks. 
Overall, 3DVLP effectively addresses the challenges in semantic 3D scene understanding by proposing these novel proxy tasks that enable effective point-cloud and language information interaction. By introducing OID, OCC, and OSC, our method can achieve state-of-the-art performance on multiple 3D vision-language multimodal tasks.
The strong generalization capabilities and short training time for fine-tuning of 3DVLP makes it suitable for a wide range of applications and multiple tasks.

The contributions of this study are summarized as follows:
(1) A 3D vision-language pre-training framework called 3DVLP has been proposed, achieving the unification of the tasks in semantic 3D scene understanding.
(2) We introduce Object-level IoU-guided Detection loss into the pre-training pipeline to obtain high-quality bounding boxes for downstream tasks. We also present two proxy tasks at the object level, including the Object-level Cross-Contrastive alignment task and Object-level Self-Contrastive learning task, which facilitate cross-modal alignment and help the model distinguish objects more accurately, respectively.
(3) We conduct extensive experiments and empirically demonstrate the effectiveness of our method in semantic 3D scene understanding.

\section{Related Work}
\subsection{Vision-language Pre-training}
Vision-language pre-training are proposed to improve the performance in downstream tasks and has been widely explored in recent approaches \cite{li2021align,chen2020uniter,radford2021learning,li2022blip,li2023blip,su2019vl,li2019visualbert,li2020oscar}. It is a common practice to pre-train the model with large-scale image-text pair datasets, usually craweled from the web \cite{radford2021learning,jia2021scaling}.
Borrowed from the insight in NLP tasks \cite{brown2020language,devlin2018bert,lan2019albert,liu2019roberta}, various learning objectives are proposed for cross-modal pre-training, enabling the model to capture the relationship between data from different modalities.
CLIP \cite{radford2021learning} aligns the unimodal image representation and language representation by contrastive loss and maximizes similarity of correct pairs.
ALBEF \cite{li2021align} and Uniter \cite{chen2020uniter} further apply image-text matching and masked language modeling tasks, enabling model to capture more complex interactions between image and text.
Li et al. introduces captioning loss in BLIP \cite{li2022blip} to address the issue of noisy image-text pairs, and further bootstraps representation learning from frozen pre-trained unimodal models in BLIP-2 \cite{li2023blip}.

% Despite success of pre-training in 2D vision language tasks, there has been no such attempt in 3D vision-language multimodal learning. Most existing work focus on task-specific end-to-end optimization and fail to generalize well in semantic 3D scene understanding. 

Pre-training for 3D vision language tasks also suffers from misaligned data across different modalities, leading to difficulties in training the fusion layer \cite{yang2022vision,li2021align}. Motivated by the common practice in 2D vision language tasks \cite{li2021align,zhang2021cross}, we introduce contrastive alignment task into 3D vision-language learning and enhance the performance of the pre-training model.

\begin{figure*}[h]
    \centering
    \includegraphics[width=\textwidth]{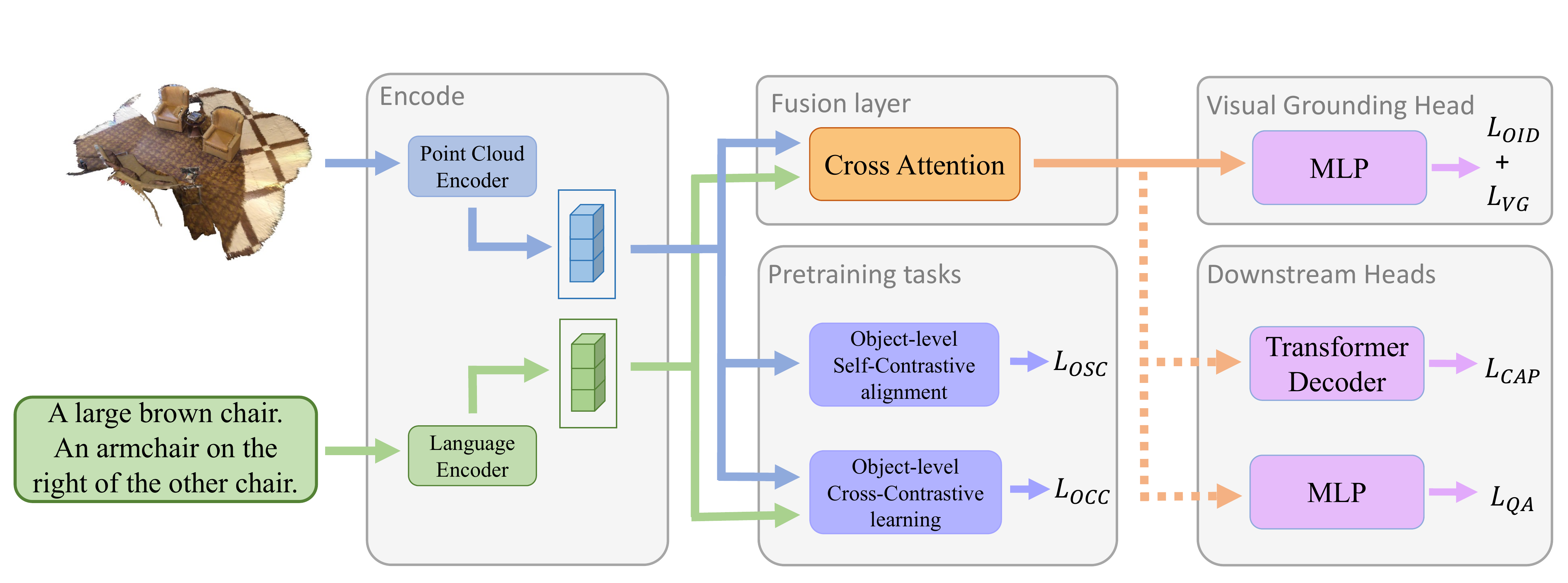}
    \caption{Pipeline of 3DVLP in semantic 3D scene understanding. 3DVLP takes visual grounding as the proxy task and utilizes Object-level IoU-guided Detection (OID) loss to boost the performance of the object detector. We also introduce Object-level Cross-Contrastive alignment task and Object-level Self-Contrastive learning task in the pre-training stage, which facilitate cross-modal alignment and enable the model to distinguish objects more accurately, respectively. }
    \label{model}
\end{figure*}
\subsection{3D Visual-Langauge Tasks}
Recently, semantic 3D scene understanding has raised great interest and has been widely explored in recent approaches across various tasks, including 3D visual grounding \cite{chen2020scanrefer,zhao20213Dvg,cai20223Djcg}, 3D dense captioning \cite{chen2021scan2cap}, and 3D question answering \cite{azuma2022scanqa}.

3D visual grounding aims to locate a region of interest in a scene based on a referring description. Chen et al. \cite{chen2020scanrefer} introduces the ScanRefer dataset and proposes an end-to-end visual grounding framework. Achlioptas et al. \cite{achlioptas2020referit3D} collects two datasets containing Nr3D and Sr3D with high-quality referential utterances.
Most existing methods rely on a detection-then-match pipeline to tackle the grounding task and aim to develop model's ability to capture the connections between proposal and  language description, which is usually implemented by a cross-attention module \cite{vaswani2017attention}.
% In a traditional way, a specific cross-attention mechanism is carefully designed in the feature fusion module to balance the mixture of text generation capacity and visual information.
For instance, 3DVG-Transformer \cite{zhao20213Dvg} introduces coordinate-guided contextual aggregation module to enhance proposal generation and cross-modal proposal disambiguation.
HAM\cite{chen2022ham} shifts attention to contextual information and develops both local and global attention module for better end-to-end grounding, while
BUTD-DETR\cite{jain2022bottom} presents a DETR-like \cite{zhu2020deformable} referential grounding model that incorporates guidance from language, points, and objects.
3D-SPS\cite{luo20223D}, however, propose the first one-stage end-to-end framework via keypoints selection and mines the cross-modal relationship based on points.

Dense captioning in 3D scene requires model to derive high-quality object bounding box and the corresponding descriptions from point cloud data.
%  and previous approaches have made significant progress in addressing this challenge.
Scan2Cap \cite{chen2021scan2cap} extends the dense captioning task to 3D scenes based on the ScanRefer dataset and establishes a messege-passing network to model the connections between objects.
SpaCap3D\cite{wang2022spatiality} investigates the relative spatiality of objects and build a spatiality-guided transformer to generate captions. Importantly, it designs a object-centric decoder by using a vision token as information carrier of the target object.

3D visual question answering is another vision-language task in which model are expected to generate a correct answer provided with the point cloud and a question. ScanQA\cite{azuma2022scanqa} collects 41k question-answer pairs and brings the question-answering task into 3D scenes. Besides, it propose a 3D-QA baseline model by casting the answer generation task as a classification problem. FE-3DGQA\cite{zhao2022towards} proposes anthoer datasets and predicts the answer through a token encoding and fusion module based on attention.

Some previous works have made efforts to capture the connection among the tasks above and dig out the basic relationship between object proposals and language expressions.
3DJCG\cite{cai20223Djcg} and D3Net \cite{chen2021d3net} model the joint training of 3D dense captioning and 3D visual grounding, thereby boosting the performance of model in both tasks. However, to the best of our knowledge, no framework has leveraged the 3D vision-language pre-training model to improve the performance of downstream tasks.
Motivated by the shared nature across different tasks in semantic 3D scene understanding, we summarize the characteristics of a pre-training model and design corresponding proxy tasks to achieve these objectives.
% Our method learns general point cloud and language encoders through pre-training over the proxy tasks and demonstrates that all 3D vision-language tasks can benefit from them. 
% The new state-of-the-art results on all the downstream tasks demonstrates the effectiveness of our framework.

\section{Method}
% \subsection{Overview}
As illustrated in Figure \ref{model}, 3DVLP first encodes point cloud and language data and further applies a cross-attention module to obtain fusion feature for downstream tasks.
The training of 3DVLP can be mainly divided into the pre-training stage and the fine-tuning stage. In the pre-training stage, 3DVLP utilizes visual grounding as the proxy task and employs Object-level IoU-guided Detection (OID) loss for high-quality object detection. Additionally, 3DVLP is pre-trained on other designed proxy tasks, including Object-level Cross-Contrastive alignment (OCC) and Object-level Self-Contrastive learning (OSC). In the finetuning stage, we transfer the backbone of 3DVLP to downstream tasks with task-specific heads.
% In this section, we first introduce Object-level IoU-guided Detection (OID) loss in 3DVLP. Then we present the designed pre-training proxy tasks including Object-level Cross-Contrastive alignment (OCC) task and Object-level Self-Contrastive learning (OSC) task.

\subsection{Object-level IoU-guided Detection Loss}
We consider visual grounding as the proxy task since it shares the same objective with the pre-training model of obtaining high-quality proposals. Additionally, we propose  Object-level IoU-guided Detection loss to enhance the performance of the object detector, as demonstrated in Fig. \ref{fig_OID}.

Specifically, we introduce the Distance IoU (DIoU) loss \cite{zheng2020distance} into the visual grounding pipeline for bounding box regression. Given the predicted proposal $\mathbf{b}_p$ and ground truth $\mathbf{b}_{gt}$, we calulate the IoU between them and have the following regression loss:

\begin{equation}
    \label{diou}
    \mathcal{L}_{D I o U}(\mathbf{b}_p,\mathbf{b}_{g t})=1-I o U+\frac{\rho^2\left(\mathbf{b}_p, \mathbf{b}_{g t}\right)}{c^2},
\end{equation}
where $c$ is the diagonal length of the smallest enclosing box covering the two boxes.
However, previous approaches\cite{zhao20213Dvg,cai20223Djcg} treats the matching stage in visual grounding task as a classification problem and use the proposal with the highest IoU as a supervised label to train the fusion module. In this case, the DIoU loss can only be applied to a single proposal, which weakens its efforts in optimization. Additionally, due to the large number of proposals generated by the detector, there can be multiple boxes pointing to the target object, and these boxes may share similar semantic information, making it difficult to achieve accurate matching with a one-hot label.

\begin{figure}[tbp]
    \centering
    \includegraphics[width=.48\textwidth]{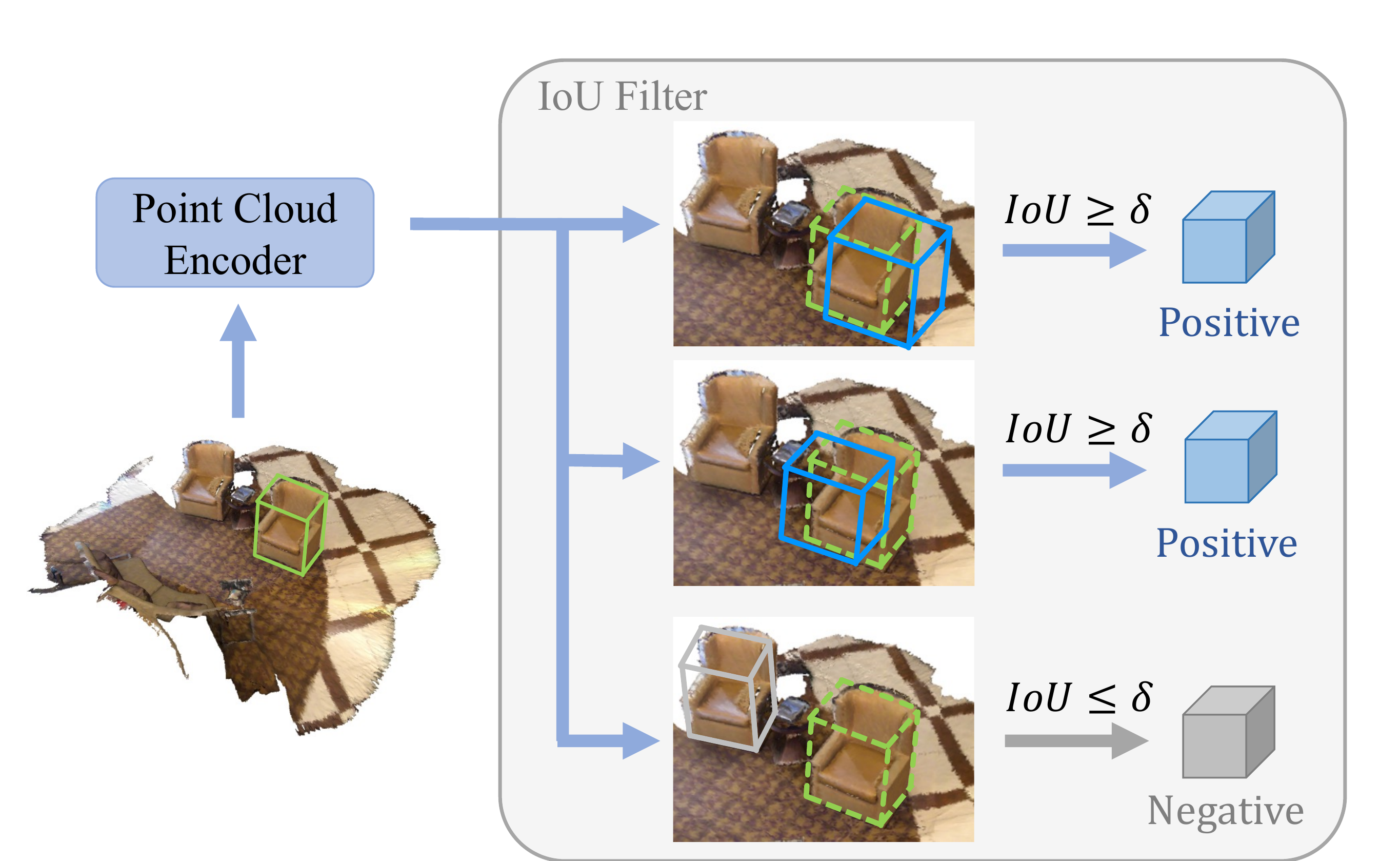}
    \caption{Illustration of the IoU filter in 3DVLP. To apply label smoothing and contrastive loss at the object level, proposals with IoU higher than a threshold $\delta$ are considered positive samples while others are regarded as the negative ones.}
    \label{filter}
\end{figure}

Label smoothing is a regularization technique that prevents the model from overconfident prediction \cite{muller2019does} and is suitable for addressing such matching problems.
Specifically, we apply label smoothing by incorporating an IoU filter into training, as shown in Fig. \ref{filter}. Given a pre-defined IoU threshold $\delta$ and the weight factor $\varepsilon$, positive proposals are filtered according to their IoU with the ground truth, and weights are assigned to them based on their total count, denoted by $K$. The weight of proposal $p$ in the soft label is shown in Equ. (\ref{smoothing}).

\begin{equation}
    \label{smoothing}
    \begin{aligned}
        y_{p} = \begin{cases}1-\varepsilon         & \text { if } IoU_p = IoU_{max}                                  \\
             \frac{\varepsilon}{K} & \text { if } IoU_p \ge \delta \text{ and } IoU_p \neq IoU_{max} \\
             0                     & \text { otherwise }
                \end{cases}
    \end{aligned}
\end{equation}
We further combine DIoU loss and label smoothing to obtain our OID loss, as demonstrated in Equ. (\ref{OID}).
\begin{equation}
    \label{OID}
    \mathcal{L}_{OID}=\sum_{p} y_p \cdot \mathcal{L}_{D I o U}(\mathbf{b}_p,\mathbf{b}_{g t}).
\end{equation}

% We found that label smoothing speed up the convergence of model while it may introduce deterioration of performance in the late stage of training. Therefore, we only use the label smoothing in early training and reuse one-hot label for high-accuracy matching after 100 epochs.

\subsection{Object-level Cross-contrastive Alignment}
As a common practice \cite{zhao20213Dvg,cai20223Djcg}, a cross-modal attention module is applied in semantic 3D scene understanding for feature fusion between language and point cloud embedding. However, it is observed that the data distribution across different modalities is not well-aligned, resulting in insufficient interaction between the embedding of proposals and the language feature.
To address this issue, contrastive learning can provide insights for embedding alignment across different distributions.  However, naive implementation over proposals is not effective, as semantically similar information from the boxes pointing at the target object conflicts with the optimization target of contrastive loss. This can ultimately lead to a deterioration in performance or even failure to converge.

% In previous research, attempts have been made to use contrastive loss to map the data of two different modalities into the same feature space for modeling. For example, in the interaction between the image modality and the text modality, many works, such as ALBEF (Li et al., 2021) and CLIP (Radford et al., 2021), align the features of images and texts using contrastive loss. Under the role of contrastive loss, the matched feature of the image-text pairs can be as similar as possible, while the unmatched feature is pulled away. These contrastive losses all compare an image sample as a whole with the text, and there is no mutual correlation between images.

\begin{figure*}[tbp]
    \centering
    \subfloat[Object-level IoU-guided Detection]{
        \includegraphics[width=.33\textwidth]{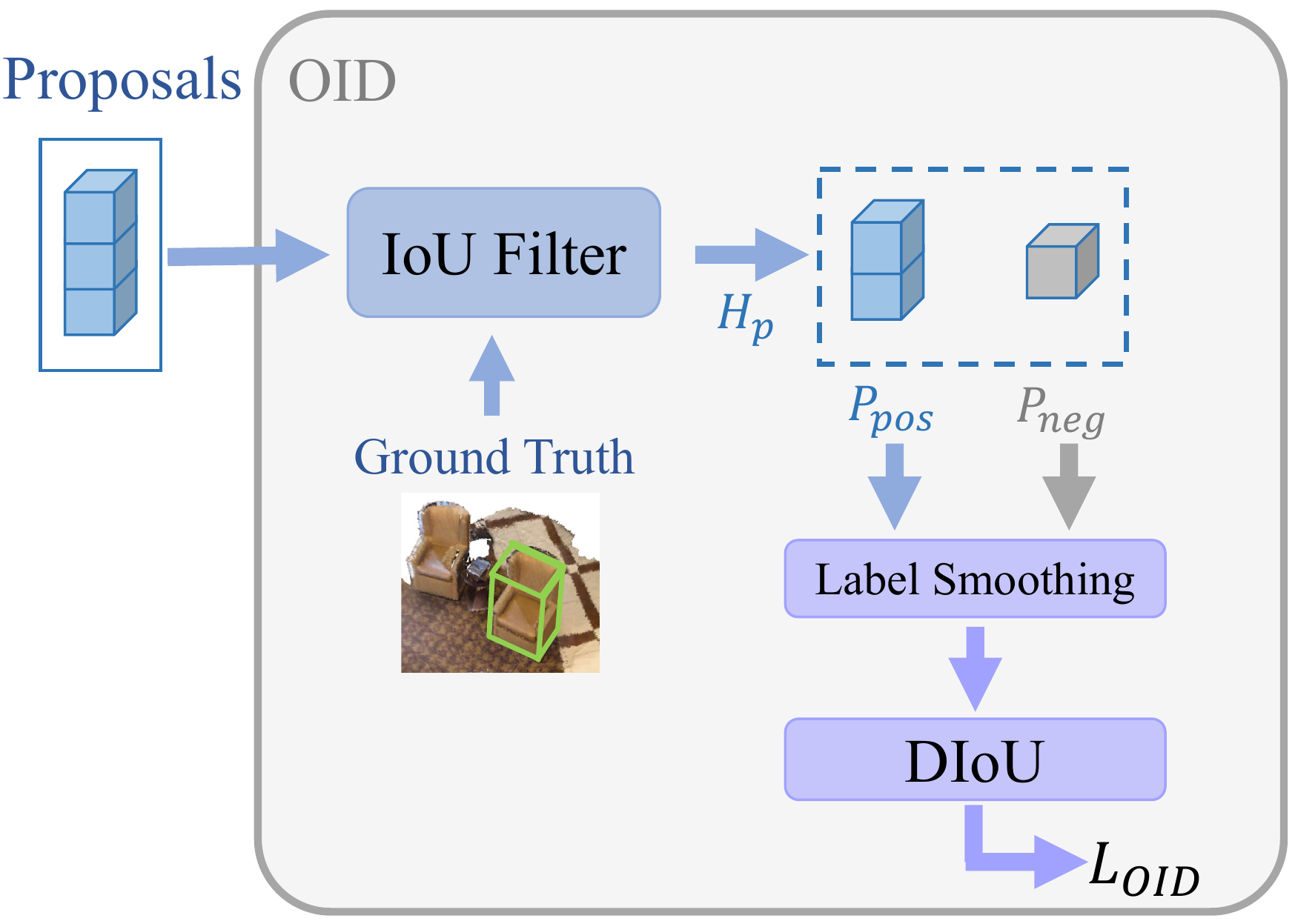}
        \label{fig_OID}
    }
    \subfloat[Object-level Cross-contrastive Alignment]{
        \includegraphics[width=.33\textwidth]{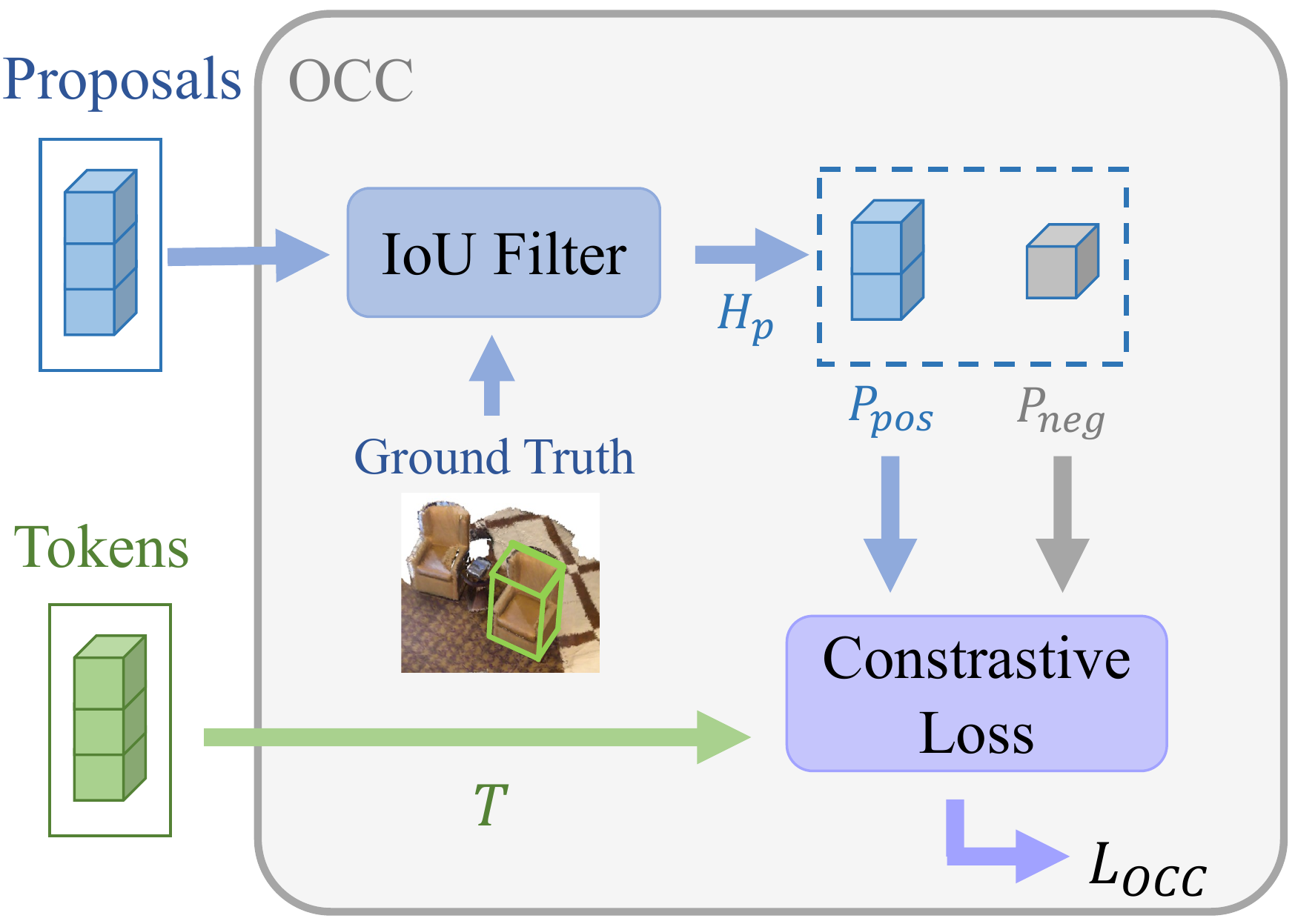}
        \label{fig_OCC}
    }
    \subfloat[Object-level Self-Contrastive Learning]{
        \includegraphics[width=.33\textwidth]{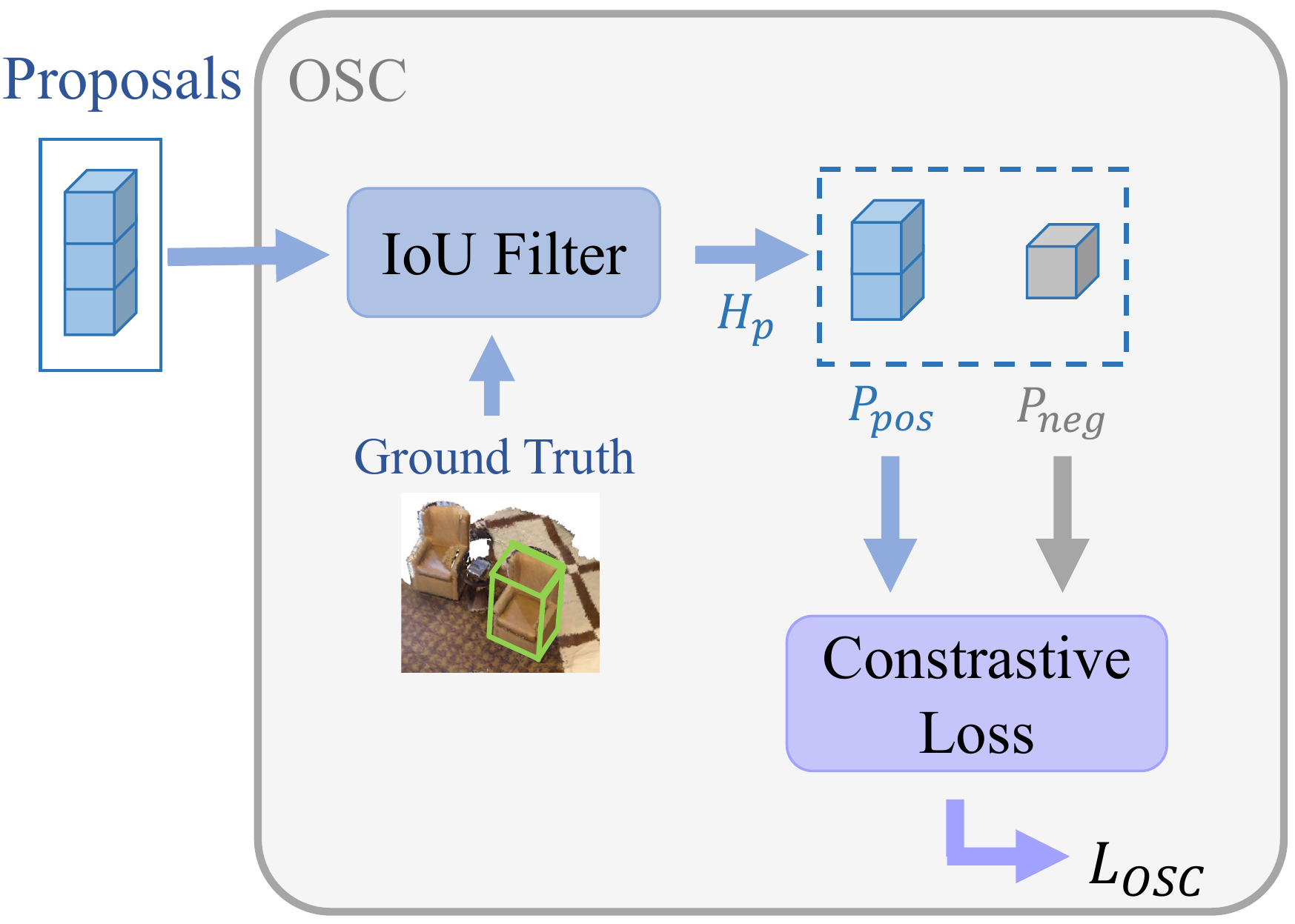}
        \label{fig_OSC}
    }
    \caption{Illustration of Object-level IoU-guided Detection (OID) loss, Object-level Cross-contrastive alignment (OCC) and Object-level Self-Contrastive learning (OSC) pre-training tasks. All the modules utilize a IoU filter to select positive proposals.}
\end{figure*}

Based on these observations, we reconsider contrastive learning at the object level and introduce the Object-level Cross-Contrastive alignment (OCC) task to enhance the performance of the cross fusion module, as shown in Fig. \ref{fig_OCC}.
The OCC task is proposed to align the distribution of cross-modal data. Specifically, in the training stage, we introduce the target detection boxes of real objects and select all the predicted boxes with IoU greater than a pre-defined threshold as positive samples since they semantically point to the target object and should have similar features. The remaining predicted boxes are considered negative samples, representing the proposals of other objects or background. We then align the features of positive samples with the language embedding and push the features of negative samples away with the contrastive loss  to achieve better cross-modal understanding.

Formally, we have the following contrastive loss, which serves as the loss function for our OCC task.

% \begin{equation}
%     \label{NCE}
%     \mathcal{L}_{\mathrm{NCE}}=-\mathbb{E}_{P(H_p, H_t)}\left[\log \frac{\exp (s(H_p, H_t))\mathbb I[IoU(\mathbf{b}_p, \mathbf{b}_{g t})>\delta_{OCC}]}{\sum_{\hat{p} \in \hat{P}} \exp (s(H_{\hat{p}},H_t))}\right],
% \end{equation}
% where $H_p$ represents the predicted box features generated by the point cloud encoder, and $H_t$ denotes the language embedding. Besides, $\mathbb I$ is the indicator function, $IoU(\cdot,\cdot)$ represents the IoU score for measuring the overlap between two object boxes, and $\delta_{OCC}$ is the IoU threshold set in the OCC task. $s(\cdot,\cdot)$ represents the similarity score function for measuring the similarity between two types of features, such as by performing a dot product operation.

% We further rewrite Equ. \ref{NCE} in a symmetrical form to obtain Equ. \ref{OCC_loss}, which serves as the contrastive loss function for our OCC task.

% \begin{equation}
%     \label{OCC_loss}
%     \begin{aligned}
%         \mathcal{L}_{\mathrm{OCC}} =-\frac{1}{2} \mathbb{E}_{P(H_p, H_t)} & \Big[  \log \frac{\exp (s(H_p, H_t))\mathbb I[IoU(\mathbf{b}_p, \mathbf{b}_{g t})\ge\delta]}{\sum_{\hat{p} \in \hat{P}} \exp (s(H_{\hat{p}},H_t))} \\  +&\log \frac{\exp (s(H_t, H_p))\mathbb I(IoU(\mathbf{b}_p, \mathbf{b}_{g t})\ge\delta)}{\sum_{\hat{p} \in \hat{P}} \exp (s(H_t,H_{\hat{p}}))}\Big].
%     \end{aligned}
% \end{equation}

\begin{equation}
    \label{OCC_loss}
    \begin{aligned}
        \mathcal{L}_{\mathrm{OCC}} =-\frac{1}{2} \mathbb{E}_{(\mathbf{b_{gt}},T)\sim D} & \Big[  \log \frac{\sum_{{p} \in {P_{pos}}}exp (s(H_p, T))}{\sum_{\hat{p} \in {P_{pos}\cup P_{neg}}} \exp (s(H_{\hat{p}},T))} \\  +&\log \frac{\sum_{{p} \in {P_{pos}}}\exp (s(T, H_p))}{\sum_{\hat{p} \in {P_{pos}\cup P_{neg}}} \exp (s(T,H_{\hat{p}}))}\Big].
    \end{aligned}
\end{equation}
% where $H_p$ represents the predicted box features generated by the point cloud encoder, and $H_t$ denotes the language embedding. $\hat{P}$ represents the proposals set containing positive samples and the negative ones. Besides, $\mathbb I$ is the indicator function, $IoU(\cdot,\cdot)$ represents the IoU score for measuring the overlap between two object boxes, and $\delta$ is the IoU threshold. $s(\cdot,\cdot)$ represents the similarity score function for measuring the similarity between two types of features, such as by performing a dot product operation.
where $H_p$ represents the embedding of proposal $p$, and $T$ denotes the language embedding. Given $\mathbb I$ as the indicator function, $IoU(\cdot,\cdot)$ as the IoU score between two boxes, and $\delta$ as the IoU threshold, we have $P_{pos}=\{p|IoU(\mathbf{b}_{p},\mathbf{b}_{g t})\ge\delta\}$ as the set of proposals containing positive samples while $P_{neg}=\{p|IoU(\mathbf{b}_{p},\mathbf{b}_{g t})<\delta\}$ containing the negative ones. $s(\cdot,\cdot)$ represents the similarity score function for measuring the similarity between two types of features, such as by performing a dot product operation.

Note that the threshold $\delta$ determines how close  positive samples should be to align with the language embedding. Specifically, when $\delta = IoU_{max}$, Equ. (\ref{OCC_loss}) only considers the proposal with the highest IoU to be the positive sample and reverts to the original formula of traditional pairwise contrastive loss.

\subsection{Object-level Self-contrastive Learning}
In semantic 3D scene understanding, the presence of similar  objects in the scene can significantly affect the model's matching performance. Therefore, a well-designed pre-training model should be capable of accurately distinguishing between objects in the scene and understanding what makes them similar or different. Achieving this is a fundamental task that challenges the model's overall understanding of the scene.
To address this issue, one effective approach is to utilize contrastive loss that incentivizes the model to capture features that differentiate objects. This can lead to an improved matching performance and enhance the model's ability to identify the target object based on the given description. Similarly, we require an object-level self-contrastive loss instead of the pairwise loss to effectively differentiate between objects and improve the model's semantic understanding of the scene.

Therefore, we introduce the Object-level Self-Contrastive learning (OSC) task for object detection, as shown in Fig. \ref{fig_OSC}. The OSC task is proposed for unimodal point cloud data and aims to optimize the embedding generated by the point cloud encoder. Based on the idea in OCC task, we utilize the IoU threshold to select positive samples and  negative ones for self contrastive learning.  By optimizing the self-contrastive loss, 3DVLP encourages the features of the boxes targeting the ground truth object to be as dissimilar as possible from those of other boxes,  thereby enabling the fusion module to distinguish different objects easily.

Following Equ. (\ref{OCC_loss}), we replace the language embedding with the embedding of proposals to obtain the corresponding contrastive loss for OSC module, as shown in Equ. (\ref{OSC_loss}).
% \begin{equation}
%     \label{OSC_loss}
%     \begin{aligned}
%          & \mathcal{L}_{\mathrm{OSC}} =-\mathbb{E}  _{P(H_p)} \\&\Big[  \log \frac{\sum_{\hat{p} \in \hat{P}}\exp (s(H_p, H_{\hat{p}}))\mathbb I[IoU(\mathbf{b}_{{p}}, \mathbf{b}_{g t})\ge\delta]\mathbb I[IoU(\mathbf{b}_{\hat{p}}, \mathbf{b}_{g t})\ge\delta]}{\sum_{\hat{p} \in \hat{P}} \exp (s(H_p,H_{\hat{p}}))}\Big ].
%     \end{aligned}
% \end{equation}
\begin{equation}
    \label{OSC_loss}
    \begin{aligned}
        \mathcal{L}_{\mathrm{OSC}} =-\mathbb{E}  _{\mathbf{b}_{gt}\sim D} \Big[  \log \frac{\sum_{{p,\hat{p}} \in {P_{pos}}}\exp (s(H_p, H_{\hat{p}}))}{\sum_{{p,\hat{p}} \in {P_{pos}\cup P_{neg}}} \exp (s(H_p,H_{\hat{p}}))}\Big ].
    \end{aligned}
\end{equation}

\subsection{Heads for Downstream Tasks}
\subsubsection{3D Visual Grounding}
3D visual grounding task involves matching a language description to the corresponding detection box in a given point cloud data of the scene.
As a common practice, we model this matching task as a classification problem by directly using the proposal features obtained from the cross-modal attention module, transforming it into a $n$-class classification task, where $n$ represents the total number of predicted boxes.
The classification label serves as the supervision information to optimize the MLP matching module using cross-entropy loss:
\begin{equation}
    \mathcal{L}_{VG}=-\frac{1}{|P_{m}|}\sum_{p_{m}\in P_{m}}y_{m}\cdot log(p_{m}),
    % \mathcal{L}_{VG}=\text{CrossEntropy}(p_{match},g_{match}),
\end{equation}
where $|P_{m}|$ denotes the total number of the proposals, $p_{m}$ represents the matching score calculated for each proposal, and $y_{m}$ represents the corresponding weight in the classification label.
% In particular, we utilize the soft label generated by label smoothing as the classification label in early training and reuse the one-hot label for high-accuracy matching in late training.

\subsubsection{3D Dense Captioning}
3D dense captioning task involves generating corresponding descriptions for all objects in a given scene. To implement the captioning module, we follow the design in 
SpaCap3D \cite{wang2022spatiality} and insert a special visual token with proposal embedding into the initial position of the sequence, which interacts with the word tokens in the attention module.
% This architecture can efficiently implement the fusion of point cloud embedding and language feature in a concise way, while meeting the requirements of the captioning task. 
We can then divide this task into training and inference stages.

In the training stage, as we already have specific information about the ground truth, we associate each real object with the nearest proposal and then use the corresponding embedding to perform captioning.
We use the natural description as the supervised label to optimize the captioning module through cross-entropy loss:
% Therefore, the loss function for the 3D dense captioning task is as follows:
\begin{equation}
    \mathcal{L}_{CAP}=-\frac{1}{|P_{cap}|}\sum_{p_{cap}\in P_{cap}}y_{cap}\cdot log(p_{cap}),
    % \mathcal{L}_{CAP}=\text{CrossEntropy}(P_{cap},G_{cap}),
\end{equation}
where $P_{cap}$ represents the score vector of each word in the sequence, while $p_{cap}$ and $y_{cap}$ denote the prediction vector and the ground truth label of a single word,respectively. Note that we also utilizes masked language modeling \cite{devlin2018bert} in dense captioning.

In the inference stage, we need to perform captioning on all the objects in the scene. Therefore, all the proposals obtained from the point cloud encoder are fed into the Non-Maximum Suppression filter and then into the captioning module as queries.

\subsubsection{3D Question Answering}
3D question answering task involves providing answers to questions about objects  given the scene data.
Following ScanQA \cite{azuma2022scanqa}, we simplify this task into a multi-class classification task for all possible answers. We count and deduplicate all answers, and consider each remaining answer as an output class in the classification task.

Specifically, a lightweight MLP is adopted to predict the score for each answer based on the fusion feature, and the answer with the highest score is selected as the final answer. Cross-entropy loss is used as the loss function to optimize the answering module:
\begin{equation}
    \label{qa_loss}
    \mathcal{L}_{QA}=-\frac{1}{|P_{qa}|}\sum_{p_{qa}\in P_{qa}}y_{qa}\cdot log(p_{qa}),
\end{equation}
where $p_{qa}$ represents the answer score computed by the model, and $y_{qa}$ represents the ground truth label.

% \begin{figure}[tbp]
%     \centering
%     \includegraphics[width=.5\textwidth]{pdf/OMLM.pdf}
%     \caption{OMLM Module in 3DVLP.}
%     \label{OMLM}
% \end{figure}

\section{Experiment}
\subsection{Datasets and Implementation Details}
\textbf{Visual Grounding Dataset:}
We select the benchmark dataset ScanRefer \cite{chen2020scanrefer} for visual grounding task. It consists of 800 3D scenes from the ScanNet dataset \cite{dai2017scannet}, each annotated with bounding boxes around objects of interest and corresponding text descriptions.
% In total, ScanRefer contains 51583 descriptions of 11046 objects, which have been split into train/val/test sets with 36655, 9508 and 2068 samples, respectively.
To evaluate our results, we employed two evaluation metrics: IoU@0.25 and IoU@0.5, which measure the percentage of times the proposals have an IoU greater than the threshold.\\
\textbf{Dense Captioning Dataset:}
We conduct experiments on Scan2Cap dataset \cite{chen2021scan2cap} to evaluate the effectiveness of our method for the dense captioning task.
% During evaluation, we assigned each object within the scene to the proposal with the highest Intersection over Union (IoU) and fed it into the captioning module to generate representation. 
Similar to  Scan2Cap, we jointly measure the quaility of the generated model with captioning matrics including CiDEr \cite{vedantam2015cider}, BlEU-4 \cite{papineni2002bleu}, METEOR \cite{banerjee2005meteor} and ROUGE \cite{lin2004rouge}, cited as C, B-4, M and R, respectively. We combine the metrics above with an IOU threshold and adopt the m@kIoU metric:
\begin{equation}
    m@kIoU=\frac{1}{N}\sum_{i=1}^{N} m_i\cdot \mathbb I(IoU\ge k)
\end{equation}
where m represents the captioning metric, k is the threshold of IoU and $\mathbb I$ stands for the indicator function.\\
\textbf{Question Answering Dataset:}
We perform a quantitative evaluation on the question answering tasks over the ScanQA dataset \cite{azuma2022scanqa}.
The ScanQA dataset consists of 41363 questions and 32337 unique answers from 800 scenes derived from the ScanNet scenes.
% The questions cover a wide range of topics, such as object attributes, spatial relationships, and scene understanding. 
Following the evaluation methodology in \cite{azuma2022scanqa}, EM@1 and EM@10 are used as the evaluation metric. EM@K is the percentage of predictions where the top K predicted answers exactly match any of the ground-truth answers.

\subsection{Implementations Details}
We first train 3DVLP over the proposed proxy tasks including visual grounding, OCC and OSC in the pre-training stage. We then evaluate our methods on the dense captioning and question answering tasks by transferring the pre-trained model and finetuning it through tasks-specific loss.
Similar to 3DJCG\cite{cai20223Djcg}, we adopt FCOS\cite{tian2019fcos} method to generate the initial object proposals and use 8 sentences per scene in a batch. We train 200 epochs over the grounding task in the pre-training stage.
Importantly, we use VoteNet \cite{qi2019deep} as our point cloud encoder and a frozen BERT\cite{devlin2018bert} as the language encoder to avoid over-fitting on short-length sentences in ScanRefer dataset.
For captioning tasks, we use a Transformer decoder with 6 layers and 128 as the hidden size. For QA task, the hidden size of the classification layer is set to be 128 as well.
We empirically set the batch size as 8 and adopt the AdamW optimizer \cite{loshchilov2017decoupled} with the cosing learning rate decay strategy. The initial learning rate is set to be 0.002 for the detector and 5e-4 for other modules in the 3DVLP.
Codes are implemented by Pytorch and run on a Nvidia 3090 GPU.

\begin{table*}[htbp]
    \centering
    % \footnotesize
    \caption{Comparison of different methods in 3D visual grounding task. We measure the percentage of the correctly predicted bounding boxes whose IoU with the ground-truth boxes are larger than 0.25 and 0.5, respectively.}
    \begin{adjustbox}{width=.75\textwidth}%
        \label{vg_result_all}
        \begin{tabular}{c|c|c|c|c|c|c|c|c}
            \toprule
            \multirow{2}{*}{ Method }                  & \multirow{2}{*}{ Venue } & \multirow{2}{*}{ Data }     & \multicolumn{2}{|c|}{ Unique } & \multicolumn{2}{|c|}{ Multiple } & \multicolumn{2}{|c}{ Overall }                                                             \\

                                                       &                          &                             & Acc@0.25                       & Acc@0.5                          & Acc@0.25                       & Acc@0.5           & Acc@0.25          & Acc@0.5           \\
            \midrule
            ScanRefer \cite{chen2020scanrefer}         & ECCV2020                 & $3 \mathrm{D}$              & 67.64                          & 46.19                            & 32.06                          & 21.26             & 38.97             & 26.10             \\
            TGNN\cite{huang2021text}                   & AAAI2021                 & $2 \mathrm{D}$              & 68.61                          & 56.80                            & 29.84                          & 23.18             & 37.37             & 29.70             \\
            InstanceRefer \cite{yuan2021instancerefer} & ICCV2021                 & 3D                          & 77.45                          & 66.83                            & 31.27                          & 24.77             & 40.23             & 32.93             \\
            FFL-3DOG\cite{feng2021free}                & ICCV2021                 & 3D                          & -                              & 67.94                            & -                              & 25.70             & -                 & 34.01             \\
            3DVG-Transformer \cite{zhao20213Dvg}       & ICCV2021                 & $3 \mathrm{D}$              & 77.16                          & 58.47                            & 38.38                          & 28.70             & 45.90             & 34.47             \\
            3DJCG\cite{cai20223Djcg}                   & CVPR2022                 & 3D                          & 78.75                          & 61.30                            & 40.13                          & 30.08             & 47.62             & 36.14             \\
            3D-SPS\cite{luo20223D}                     & CVPR2022                 & 3D                          & 81.63                          & 64.77                            & 39.48                          & 29.61             & 47.65             & 36.43             \\
            BUTD-DETR\cite{jain2022bottom}             & ECCV2022                 & 3D                          & 84.20                          & 66.30                            & \textbf{46.60}                 & \textbf{35.10}    & \textbf{52.20}    & \underline{39.80} \\
            \midrule
            ScanRefer \cite{chen2020scanrefer}         & ECCV2020                 & $2 \mathrm{D}+3 \mathrm{D}$ & 76.33                          & 53.51                            & 32.73                          & 21.11             & 41.19             & 27.40             \\
            SAT\cite{yang2021sat}                      & ICCV2021                 & $2 \mathrm{D}+3 \mathrm{D}$ & 73.21                          & 50.83                            & 37.64                          & 25.16             & 44.54             & 30.14             \\
            3DVG-Transformer \cite{zhao20213Dvg}       & ICCV2021                 & $2 \mathrm{D}+3 \mathrm{D}$ & 81.93                          & 60.64                            & 39.30                          & 28.42             & 47.57             & 34.67             \\
            Multi-View Trans \cite{huang2022multi}     & CVPR2022                 & $2 \mathrm{D}+3 \mathrm{D}$ & 77.67                          & 66.45                            & 31.92                          & 25.26             & 40.80             & 33.26             \\
            3D-SPS\cite{luo20223D}                     & CVPR2022                 & $2 \mathrm{D}+3 \mathrm{D}$ & \underline{84.12}              & 66.72                            & 40.32                          & 29.82             & 48.82             & 36.98             \\
            3DJCG\cite{cai20223Djcg}                   & CVPR2022                 & $2 \mathrm{D}+3 \mathrm{D}$ & 83.47                          & 64.34                            & 41.39                          & 30.82             & 49.56             & 37.33             \\
            D3Net \cite{chen2021d3net}                 & ECCV2022                 & $2 \mathrm{D}+3 \mathrm{D}$ & -                              & \textbf{70.35}                   & -                              & 30.05             & -                 & 37.87             \\
            \midrule
            3DVLP                                      & -                        & $2 \mathrm{D}+3 \mathrm{D}$ & \textbf{85.18}                 & \underline{70.04}                & \underline{43.65}              & \underline{33.40} & \underline{51.70} & \textbf{40.51}    \\
            % 3DVLP                                      & -                        & $2 \mathrm{D}+3 \mathrm{D}$ & 83.03                          & \underline{67.78}                & \textbf{43.78}                 & \textbf{34.70} & \textbf{52.39} & \textbf{40.79}    \\
            \bottomrule
        \end{tabular}
    \end{adjustbox}
\end{table*}

\begin{table*}[tbp]
    \centering
    \caption{Comparison of different methods in 3D dense captioning task. We report the result with the percentage of the predicted bounding boxes whose IoU with the ground truth are greater than 0.25 and 0.5.}
    \footnotesize
    \label{cap_result}
    \begin{adjustbox}{width=.75\textwidth}%
        \begin{tabular}{c|c|cccc|cccl}
            \toprule
            Method                            & Venue      & C@0.25            & B-4@0.25          & M@0.25            & R@0.25            & C@0.5             & B-4@0.5           & M@0.5             & R@0.5             \\
            \midrule
            Scan2Cap \cite{chen2021scan2cap}  & CVPR 2021  & 56.82             & 34.18             & 26.29             & 55.27             & 39.08             & 23.32             & 21.97             & 44.78             \\
            MORE\cite{jiao2022more}           & ECCV 2022  & 62.91             & 36.25             & 26.75             & 56.33             & 40.94             & 22.93             & 21.66             & 44.42             \\
            SpaCap3D\cite{wang2022spatiality} & IJCAI 2022 & 63.30             & 36.46             & 26.71             & 55.71             & 44.02             & 25.26             & 22.33             & 45.36             \\
            3DJCG \cite{cai20223Djcg}         & CVPR2022   & \underline{64.70} & \underline{40.17} & \underline{27.66} & \underline{59.23} & \underline{49.48} & \underline{31.03} & 24.22             & 50.80             \\
            D3Net \cite{chen2021d3net}        & ECCV2022   & -                 & -                 & -                 & -                 & 46.07             & 30.29             & \underline{24.35} & \underline{51.67} \\
            % Vote2Cap-DETR                     & 3D             & \textbf{61.81}    & \textbf{34.46}    & 26.22             & \textbf{54.40}                                                                                    \\
            \midrule
            % 3DVLP                             & -              & \textbf{65.19}    & \underline{40.03} & \textbf{35.88}    & \textbf{59.99}    & \textbf{51.43}    & \textbf{31.86}    & \textbf{33.77}    & \textbf{51.99}    \\
            3DVLP                             & -          & \textbf{67.25}    & \textbf{41.30}    & \textbf{36.27}    & \textbf{61.53}    & \textbf{54.41}    & \textbf{34.10}    & \textbf{34.34}    & \textbf{54.28}    \\
            \bottomrule
        \end{tabular}
    \end{adjustbox}
\end{table*}
\begin{table}[tbp]
    \centering
    \footnotesize
    \caption{Comparison of different methods in 3D question answering task. The results are presented with the percentage of predictions where the top K predicted answers exactly match any of the ground-truth answers. We also report Acc@0.25 and Acc@0.5 metrics, similar to the visual grounding metrics.}
    \begin{adjustbox}{width=.45\textwidth}%
        \label{qa_result}
        \begin{tabular}{c|ccccc}
            \toprule
            Method                                  & EM@1              & EM@10             & Acc@0.25          & Acc@0.5           \\
            \midrule
            % RandomImage+MCAN (real)                & 19.19             & 48.15             & -                 & -                 \\
            % TopDownImage+Oscar(real)   & 19.38             & 46.37             & -                 & -                 \\
            VoteNet \cite{qi2019deep}+MCAN          & 17.33             & 45.54             & -                 & -                 \\
            ScanRefer \cite{chen2020scanrefer}+MCAN & 18.59             & 46.76             & 23.53             & 11.76             \\
            ScanQA\cite{azuma2022scanqa}            & 21.05             & 51.23             & 24.96             & 15.42             \\
            FE-3DGQA\cite{zhao2022towards}          & \underline{22.26} & \underline{54.51} & \underline{26.62} & \underline{18.83} \\
            \midrule
            % 3DVLP                          & -        & \textbf{24.63}    & \textbf{57.34}    & \textbf{31.51}    & \textbf{25.57}    \\
            3DVLP                                   & \textbf{24.03}    & \textbf{57.91}    & \textbf{33.38}    & \textbf{26.12}    \\
            \bottomrule
        \end{tabular}
    \end{adjustbox}
\end{table}

\subsection{Baselines}
% \begin{enumerate}
% \item \textbf{Visual Grounding:} 
In 3D visual grounding task, we compare 3DVLP with the benchmark methods including "3D" models \cite{chen2020scanrefer,huang2021text,yuan2021instancerefer,feng2021free,zhao20213Dvg,cai20223Djcg,luo20223D,jain2022bottom} and "2D+3D" models \cite{chen2020scanrefer,yang2021sat,zhao20213Dvg,huang2022multi,luo20223D,chen2021d3net} . The "3D" models only utilizes raw attributes in point cloud input features, such as the coordinates, colors, and normal vectors of the original point cloud, while "2D+3D" models use 2D multi-view features as additional inputs.
% \item 
In 3D dense captioning task, we choose end-to-end models of this task as the baseline algorithms for comparison. \cite{chen2021scan2cap,jiao2022more,wang2022spatiality,cai20223Djcg,chen2021d3net}.
% \item 
In 3D question answering task, we compare 3DVLP with ScanQA\cite{azuma2022scanqa}, FE-3DGQA\cite{zhao2022towards} and 2D models with MCAN\cite{yu2019deep}.

% \end{enumerate}

\begin{table}[tbp]
    \centering
    \caption{Ablation analysis. We provide quantitative results of the overall accuracy in visual grounding and the metric under IoU=0.5 setting in dense captioning.}
    \label{ablation}
    % \footnotesize
    \begin{adjustbox}{width=.45\textwidth}%
        \begin{tabular}{c|c|c|c|c|c|c|c|c}
            % \begin{tabular}{c|c|c|c|c}
            \toprule
            \multicolumn{3}{c|}{Module} & \multicolumn{2}{|c}{Visual Gounding} & \multicolumn{4}{|c}{Dense Captioning  }                                                                                                       \\
            % \multicolumn{3}{c|}{Module} & \multicolumn{2}{|c}{Unique}            & \multicolumn{2}{|c}{Multiple}           & \multicolumn{2}{|c}{Overall} & \multicolumn{4}{|c}{IoU=0.25  } & \multicolumn{4}{|c}{IoU=0.5}                                                                                                                            \\
            OID                         & OCC                                  & OSC                                     & Acc@0.25       & Acc@0.5        & C@0.5          & B-4@0.5        & M@0.5          & R@0.5          \\
            \midrule
                                        &                                      &                                         & 50.59          & 37.96          & 53.12          & 31.90          & 33.93          & 52.27          \\
            \midrule
            \Checkmark                  &                                      &                                         & 50.46          & 39.49          & 52.91          & 33.91          & 34.28          & 54.08          \\
                                        & \Checkmark                           &                                         & 51.15          & 38.44          & 53.24          & 32.79          & 33.98          & 52.99          \\

                                        &                                      & \Checkmark                              & 50.91          & 38.28          & 51.41          & 32.93          & 34.00          & 52.94          \\
            \midrule
            \Checkmark                  & \Checkmark                           & \Checkmark                              & \textbf{51.70} & \textbf{40.51} & \textbf{54.41} & \textbf{34.10} & \textbf{34.34} & \textbf{54.28} \\
            \bottomrule
        \end{tabular}
    \end{adjustbox}
\end{table}

\subsection{Comparison with State-of-the-art Methods}
\subsubsection{3D visual grounding task}
We present the results of 3D visual grounding in Table \ref{vg_result_all}. The results indicate that 3DVLP performs remarkably well and outperforms the baselines by a large margin.
In terms of unique scenes, 3DVLP achieves the highest accuracy in Acc@0.5 and ranks second in Acc@0.25, indicating the significant impact of our OID loss in developing the model's ability to identify high-quality bounding boxes. Previous work solely optimizes the center and size of the proposals, while the introduction of the OID loss improves the quality of proposals targeting the ground truth object.

Furthermore, when comparing multiple and unique metrics, previous works suffers from issues related to the presence of similar objects in the scene, leading to poor matching results. However, the introduction of OSC and OCC tasks in 3DVLP enables it to achieve competitive performance in multiple metrics, showcasing its ability to accurately locate objects in complex scenes. In the overall metric, 3DVLP's performance surpasses the baseline by 0.71\% in Acc@0.5 and also ranks second in Acc@0.25, demonstrating its effectiveness in 3D visual grounding.

\begin{figure}[t]
    \centering
    \subfloat{
        \includegraphics[width=0.23\textwidth]{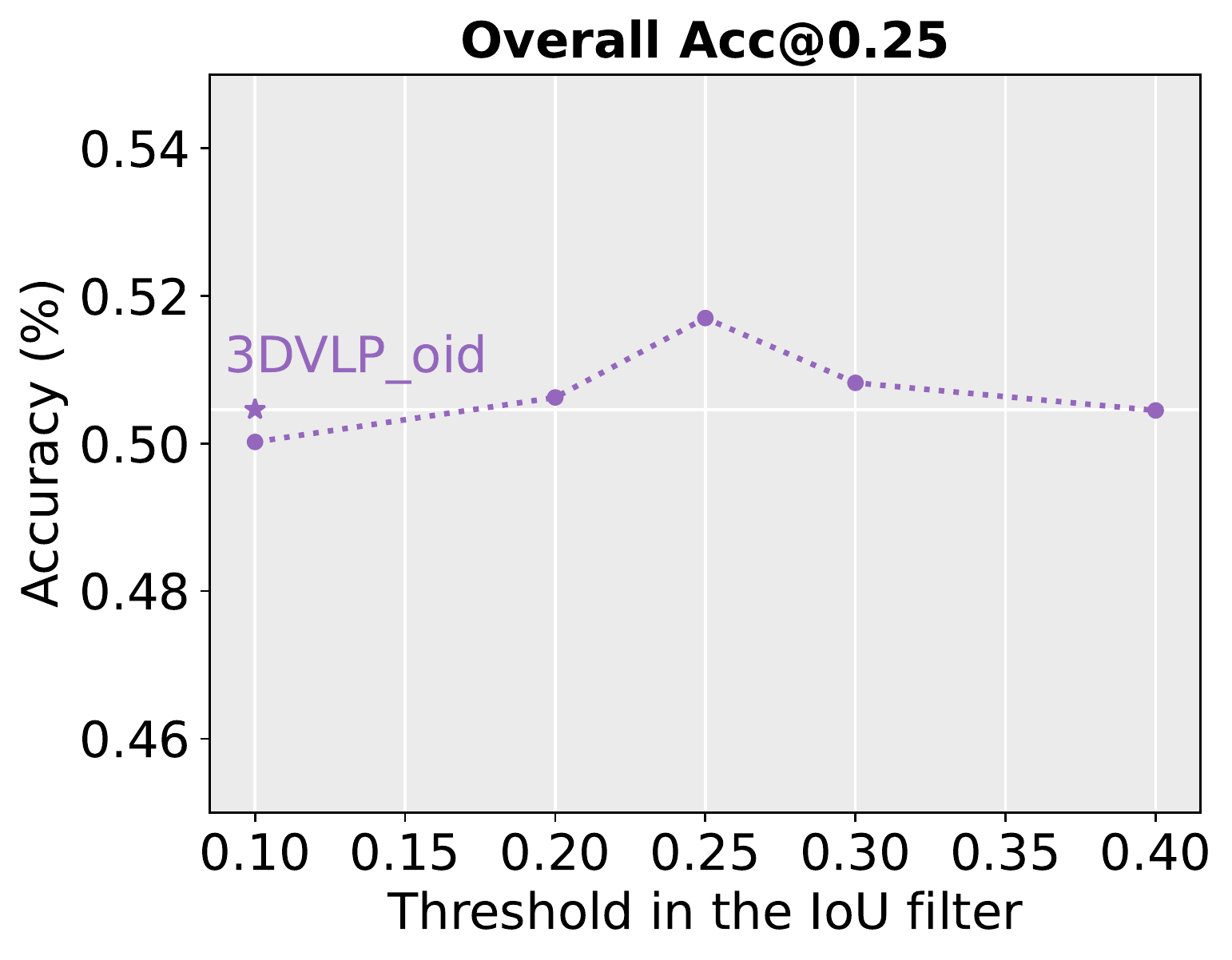}
        \label{ablation_25}
        % \label{ablation_25}
    }
    \subfloat{
        \includegraphics[width=0.23\textwidth]{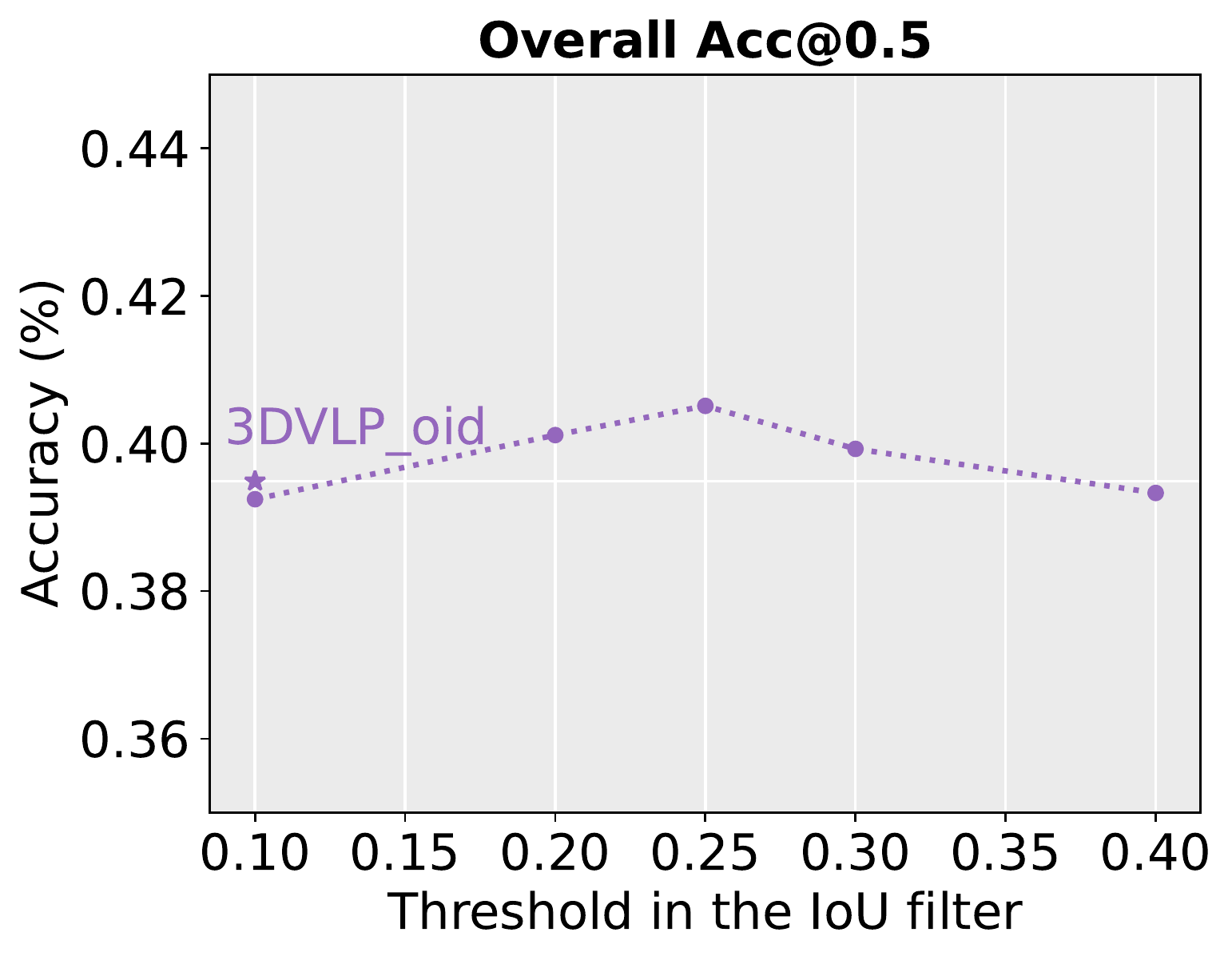}
        \label{ablation_5}
    }
    \caption{Comparison of the performance when using different threshold in the IoU filter. In addition, we compare a variant of 3DVLP with only OID loss, referred to as 3DVLP\_oid.}
    \label{threshold}
\end{figure}
% 3DVLP also achieved excellent results in the pre-training task. Next, we freeze its backbone network and train the downstream task heads, allowing it to transfer its learned high-quality point cloud and text features to other tasks, further enhancing its ability in 3D visual annotation and 3D visual question answering tasks.
\begin{figure*}[tbp]
    \centering
    \includegraphics[width=.85\textwidth, page=2]{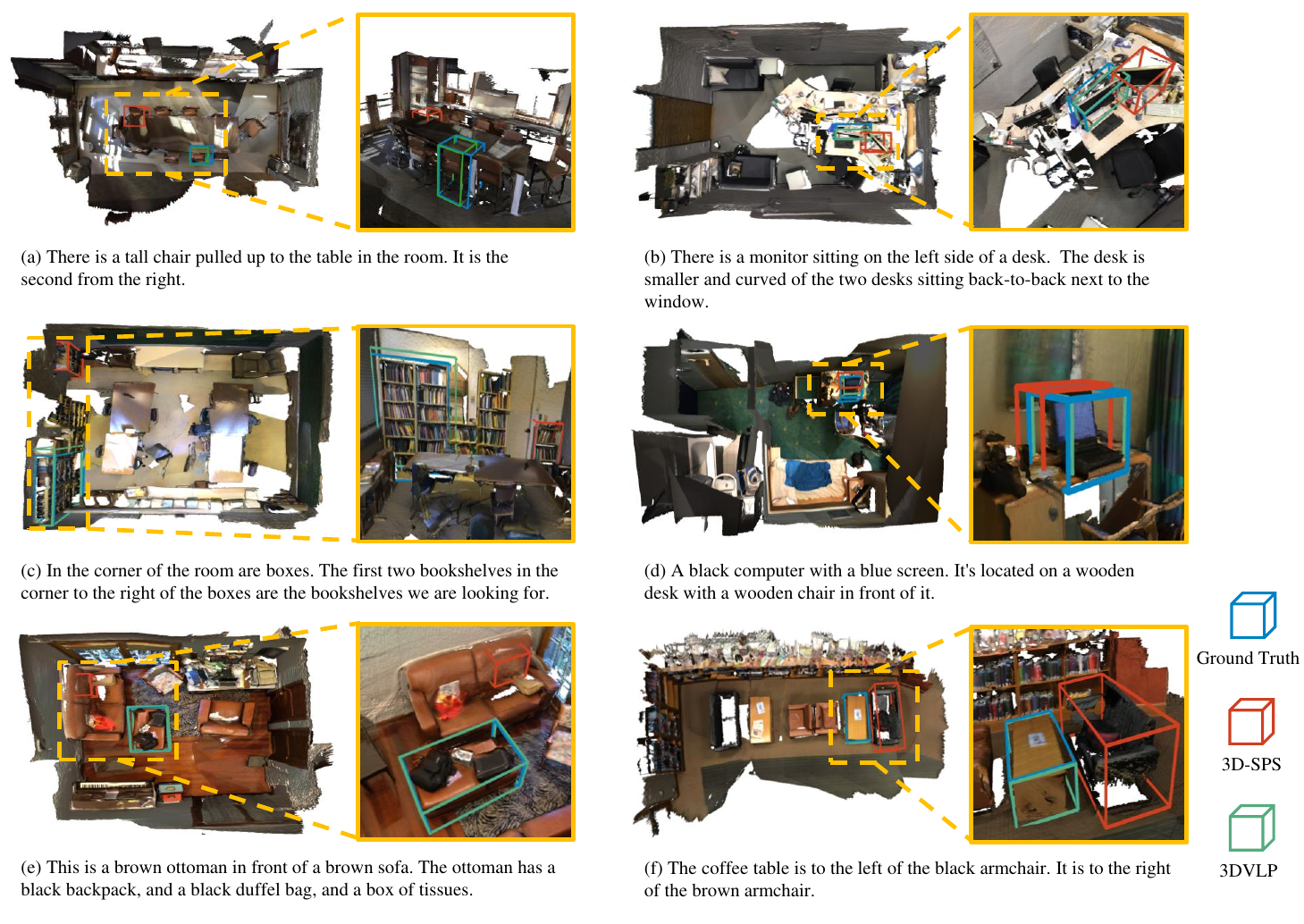}
    \caption{Qualitative results of 3DVLP and 3D-SPS. We mark the ground truth in blue, 3D-SPS in red and 3DVLP in green. }
    \label{visualization}
\end{figure*}
\subsubsection{3D dense captioning task}
As presented in Table \ref{cap_result}, it is evident that 3DVLP shows excellent transfer performance in dense captioning task. Importantly, the point cloud encoder in 3DVLP extracts universal features that generalize well in dense captioning, enabling 3DVLP to outperform other baselines by a large extent. Specifically, 3DVLP achieves a remarkable improvement of 2.55\%, 4.93\%, 2.30\%, and 2.61\% in terms of C@0.25, C@0.5, R@0.25, and R@0.5, respectively.
Moreover, the results show that 3DVLP outperforms the second baseline by 8.61\% in M@0.25 and 9.99\% in M@0.5. Among various evaluation metrics, METEOR focuses on capturing the semantic similarity and fluency between the output and the ground truth, thereby indicating the generalization ability of the encoder in 3DVLP.
In comparison to SpaCap3D, which shares the same decoder architecture as 3DVLP, we observe a significant performance boost resulting from the pre-training backbone, thus demonstrating the effectiveness of the proxy tasks designed in the pre-training stage.
\subsubsection{3D question answering task}
From the results in Table \ref{qa_result},  the most striking observation emerging from
the comparison is that 3DVLP consistently outperforms other methods and improves the performance in the question answering task.
For example, 3DVLP achieves approximately 1.7\%-2.4\% improvement in EM@1 and EM@10 compared to the baseline.
Moreover, it can be concluded that question answering benefits from the pre-training model when compared to ScanQA, as 3DVLP utilizes the same classification head.
Furthermore, 3DVLP provides a boost by 6.76\% and 7.23\% in Acc@0.25 and Acc@0.5, respectively. However, it is noteworthy that the results are lower than those achieved in visual grounding, primarily due to the inclusion of the task-specific loss in the question answering task.

\subsection{Ablation Study}
\textbf{Does the OID loss and the designed proxy tasks benefit downstream tasks?}
We conducted a series of ablation experiments to investigate the contribution of each module in 3DVLP. The results in Table \ref{ablation} demonstrate that both visual grounding and dense captioning tasks benefit from each proposed module. In visual grounding, the OID loss significantly improves the quality of the predicted bounding boxes, thereby enhancing Acc@0.5 to a large degree. Furthermore, neither the introduction of OSC nor OCC provides a remarkable boost in Acc@0.25, indicating the  superiority of modeling optimization at the object level in complex scenes.  In dense captioning, the improvement of the model is consistent with that in visual grounding by combining the modules together.
\\\textbf{Is the improvement in OSC and OCC sensitive to the threshold used the IoU filter?} To have a better understanding of the threshold $\delta$ used in the IoU filter, we estimate the results of the overall Acc in visual grounding with the varying $\delta$.
Moreover, we also include 3DVLP with only OID loss as a base variant, referred as 3DVLP\_oid.
As shown in Fig. \ref{threshold}, the performance  obviously improves when increasing the threshold from 0.1 to 0.25. This is because proposals targeting other objects can be incorrectly considered as positive samples and thus mislead the training optimization when using a low threshold. However, we further increase the threshold and observe that the improvement is not consistent. The performance drops with a large threshold since model will regard proposals that are not good enough as negative samples, resulting in semantic divergence. This is similar to what happens with the traditional pairwise contrastive loss. Therefore, based on our results, we believe that selecting a threshold of 0.25 in the IoU filter is a reasonable tradeoff.

\subsection{Qualitative Results}
To further explore how 3DVLP improves the performance in visual grounding, we provide the comparison results with 3D-SPS as shown in Figure \ref{visualization}.
Figure \ref{visualization}(d) indicates that OID loss contribute to more high-quaility bounding boxes, thereby boosting the performance.  Additionally, these examples demonstrate that 3DVLP has a better understanding of the relationship between scene and language as a result of incorporating OSC and OCC, leading to more reliable visual grounding results.

\section{Conclusion}
This paper investigates the shared nature across different tasks in semantic 3D scene understanding and proposes a contrastive 3D vision-language pre-training framework named 3DVLP, which transfers flexibly in the downstream tasks. 3DVLP introduces the object-level IoU-guided detection loss to obtain high-quaility proposals, aligns the point cloud representation and language representation by training over object-level cross-contrastive alignment task and develops its ability to distinguish different objects in the scene through object-level self-contrastive learning task, which defines a new paradigm for the 3D vision-language pre-training model.
Comprehensive experiments reveal the generalization ability and superiority of 3DVLP over all downstream tasks in semantic 3D scene understanding, leading to a new state-of-the-art performance. Future work needs to focus on dealing with the fusion of point cloud and language, desirably about the full interaction of multi-level information.
\bibliographystyle{ACM-Reference-Format}
\bibliography{sample-base}


\appendix
\clearpage
\newcommand{\appendixhead}%
{\textbf{\huge Appendix}}
\appendixhead

\section{Overview}
In Section \ref{dataset_details}, we provide more details of the datasets used in the downstream tasks.
In Section \ref{ablation_qa}, we conduct the ablation study in 3D question answering and show the effect of each module in 3DVLP.
In Section \ref{scratch}, we compare 3DVLP with variant that train from scratch to verify the effectiveness and superiority of the pre-training stage.
In Section \ref{tsne}, we provide the t-SNE \cite{van2008visualizing} visualization of proposal features in the scene from 3DVLP and variant without OID, OCC and OSC.
In Section \ref{visual}, we show more qualitative results in 3D dense captioning task.
\section{Dataset Details}\label{dataset_details}
To benchmark the performance in the downstream tasks, we select different datasets in the experiments and describe their detailed information below.

\textbf{ScanRefer}\cite{chen2020scanrefer}.
ScanRefer is a large-scale benchmark dataset designed for 3D object localization and referred object segmentation in real-world scenes. The dataset consists of textual descriptions of objects present in the scene and their corresponding 3D bounding boxes. The main objective of the dataset is to enhance the performance of 3D object detection and recognition in real-world scenarios by providing a benchmark for models that can understand natural language descriptions of objects and their spatial relationships. The dataset comprises a total of 51,583 descriptions of 11,046 objects, which have been divided into train/val/test sets with 36,655, 9,508, and 2,068 samples, respectively. Additionally, ScanRefer categorizes the data into two subsets: "unique" and "multiple." The "unique" subset contains grounding data with only a single object of its class in the scene, while the "multiple" subset contains data with more than one object of a particular class in the scene.

\textbf{Scan2Cap} \cite{chen2021scan2cap}.
Scan2Cap is a dataset designed for generating natural language descriptions of indoor scenes from 3D point cloud data. The primary objective of this dataset is to provide a benchmark for models that can generate natural language descriptions of indoor scenes using 3D point cloud data. The dataset is highly useful for evaluating the effectiveness of different techniques for combining computer vision and natural language processing to generate coherent and accurate descriptions of indoor scenes.
To simplify the problem, Scan2Cap truncates descriptions longer than 30 tokens in ScanRefer and adds two special tokens, namely SOS and EOS, to indicate the start and end of the description. Additionally, Scan2Cap follows the same data division as ScanRefer, dividing the 36,665 and 9,508 samples into train and validation sets, respectively.

\textbf{ScanQA} \cite{azuma2022scanqa}.
The ScanQA dataset is a benchmark dataset designed for visual question answering (VQA) in 3D scenes. Based on the ScanNet dataset, it provides high-quality 3D scanning data of indoor scenes with corresponding questions and answers. The dataset covers a wide range of object categories, making it a challenging benchmark for VQA models. ScanQA contains a total of 41,363 questions and 58,191 answers, including 32,337 unique questions and 16,999 unique answers. It follows the same training, validation, and test set splits as in ScanRefer.

\begin{table}[htbp]
    \centering
    \caption{Ablation analysis in question answering. We report the percentage of exactly matched predictions.}
    \label{ablation_result_qa}
    % \footnotesize
    % \begin{adjustbox}{width=.45\textwidth}%
    \begin{tabular}{c|c|c|c|c}
        % \begin{tabular}{c|c|c|c|c}
        \toprule
        \multicolumn{3}{c|}{Module} & \multicolumn{2}{|c}{Question Answering}                                                \\
        % \multicolumn{3}{c|}{Module} & \multicolumn{2}{|c}{Unique}            & \multicolumn{2}{|c}{Multiple}           & \multicolumn{2}{|c}{Overall} & \multicolumn{4}{|c}{IoU=0.25  } & \multicolumn{4}{|c}{IoU=0.5}                                                                                                                            \\
        OID                         & OCC                                     & OSC        & EM@1           & EM@10          \\
        \midrule
                                    &                                         &            & 23.23          & 56.66          \\
        \midrule
        \Checkmark                  &                                         &            & 22.58          & 55.94          \\
                                    & \Checkmark                              &            & 23.80          & 57.88          \\

                                    &                                         & \Checkmark & \textbf{24.75} & 57.24          \\
        \midrule
        \Checkmark                  & \Checkmark                              & \Checkmark & {24.03}        & \textbf{57.91} \\
        \bottomrule
    \end{tabular}
    % \end{adjustbox}
\end{table}

\section{Ablation Analysis in 3D Question Answering}\label{ablation_qa}
We conducted ablation experiments in 3D question answering and report the results in Table \ref{ablation_result_qa}. As shown in the results, the OCC and OSC modules provide a positive boost, while the OID module results in a slight drop in performance. We hypothesize that this is because adding alone the OID loss does not enable the model to handle the complex relationship in the scene according to the questions.

\begin{figure}[htbp]
    \centering
    \subfloat[3DVLP-base]{
        \includegraphics[width=0.23\textwidth]{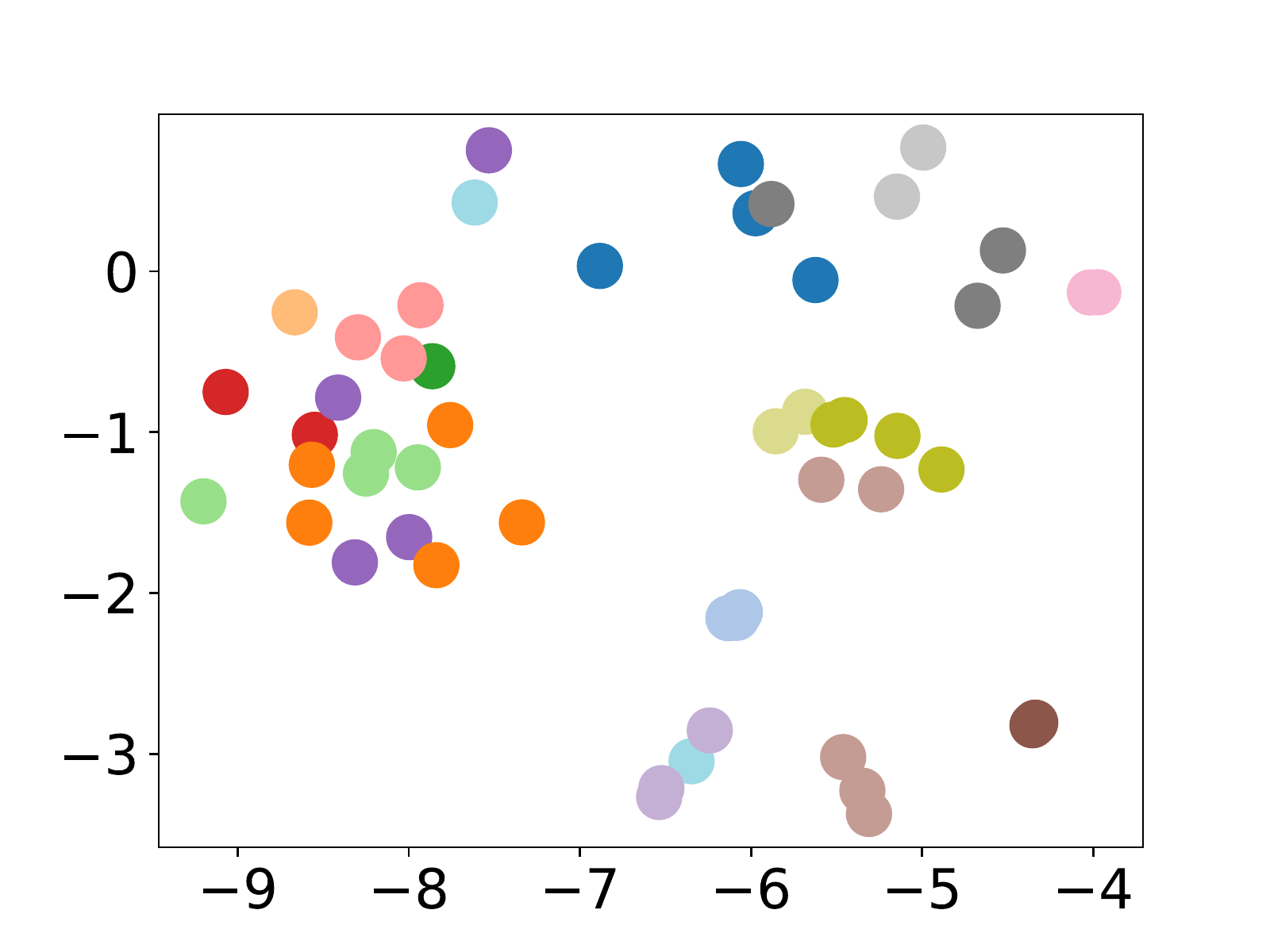}
        % \label{ablation_25}
        % \label{ablation_25}
    }
    \subfloat[3DVLP]{
        \includegraphics[width=0.23\textwidth]{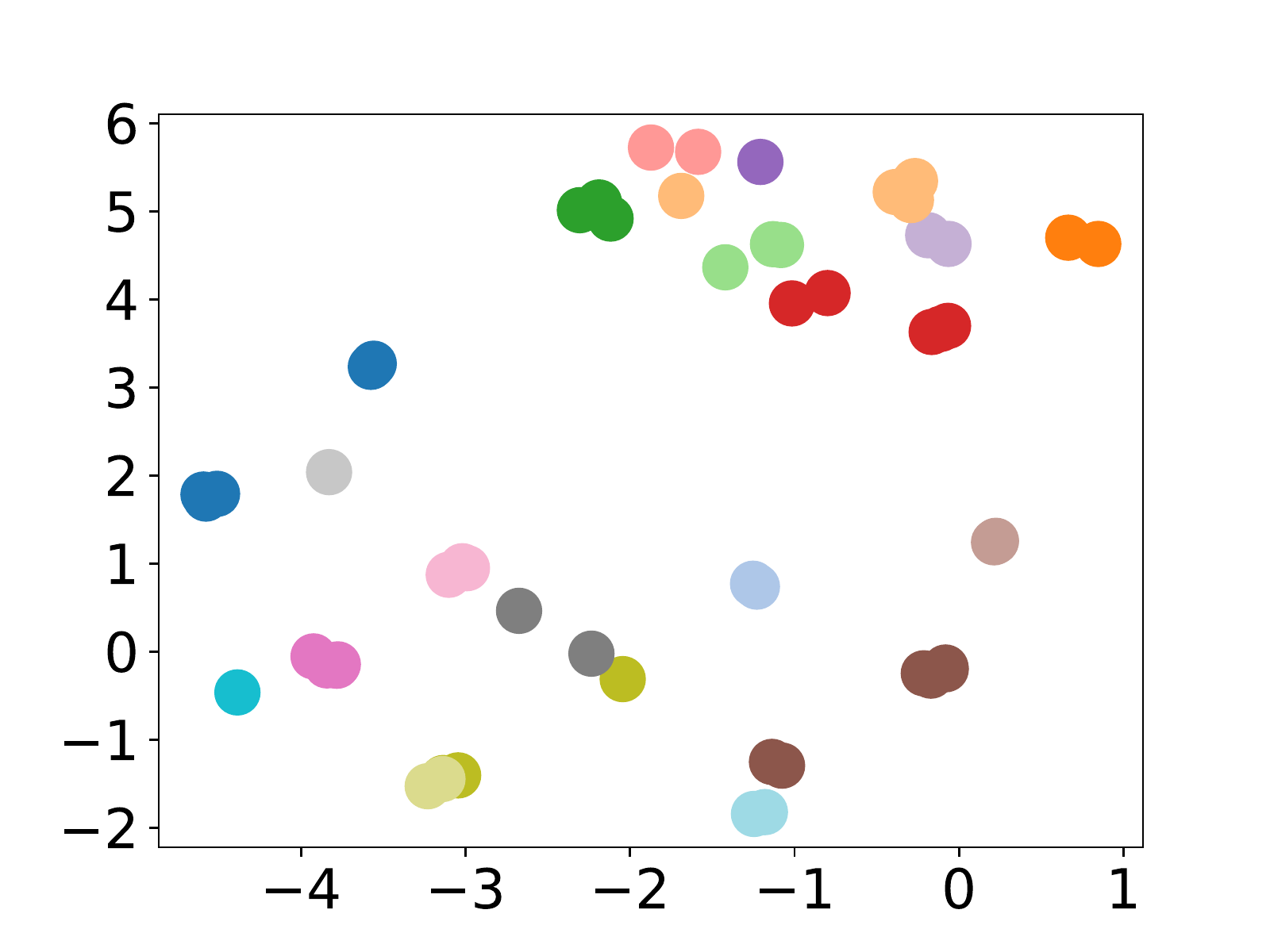}
        % \label{ablation_5}
    }
    \caption{t-SNE visualization of proposal features in the scene. 3DVLP-base is the variant of 3DVLP that does not include OID, OSC and OCC modules. }
    \label{tsne_visual}
\end{figure}

\begin{table*}[t]
    \centering
    \caption{Comparison results between 3DVLP and its variants trained from scratch. Specifically, we compare 3DVLP trained from scratch for 20 epochs (denoted as "scratch-20") and 3DVLP trained from scratch until full convergence (denoted as "scratch-full").}
    \label{result_scratch}
    \begin{adjustbox}{width=\textwidth}%
        \begin{tabular}{c|c|c|c|c|c|c|c|c|c|c}
            % \begin{tabular}{c|c|c|c|c}
            \toprule
                         & \multicolumn{8}{|c}{Dense Captioning} & \multicolumn{2}{|c}{Question Answering}                                                                                                                                         \\
            Method       & C@0.25                                & B-4@0.25                                & M@0.25         & R@0.25         & C@0.5          & B-4@0.5        & M@0.5          & R@0.5          & EM@1           & EM@10          \\
            \midrule
            scratch-20   & 59.71                                 & 38.89                                   & 35.55          & 59.57          & 41.38          & 26.70          & 32.46          & 48.01          & 21.51          & 53.99          \\
            scratch-full & 64.14                                 & 38.59                                   & 35.63          & 58.94          & 48.81          & 30.08          & 33.29          & 50.28          & 22.18          & 54.04          \\
            \midrule
            3DVLP        & \textbf{66.63}                        & \textbf{40.85}                          & \textbf{36.12} & \textbf{61.03} & \textbf{54.41} & \textbf{34.10} & \textbf{34.34} & \textbf{54.28} & \textbf{24.03} & \textbf{57.91} \\
            \bottomrule
        \end{tabular}
    \end{adjustbox}
\end{table*}

\begin{figure*}[th]
    \centering
    \includegraphics[width=.95\textwidth]{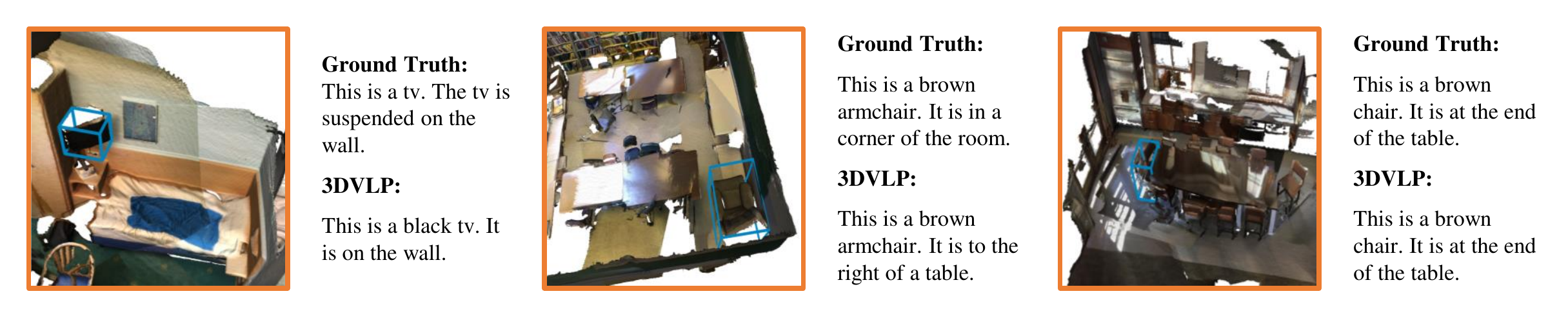}
    \caption{Qualitative results in dense captioning.}
    \label{cap}
\end{figure*}
\section{Comparison with Training from Scratch}\label{scratch}
We conduct extensive experiments and provide comparison results between 3DVLP and training from scratch in downstream tasks to evaluate the effectiveness of the pre-training stage. Since 3DVLP is fine-tuned in downstream tasks for 20 epochs, we used its variants that are trained from scratch for 20 epochs and trained from scratch until full convergence as baselines, denoted as scratch-20 and scratch-full, respectively.

As shown in Table \ref{result_scratch}, the results demonstrate that the pre-training stage over the proxy tasks provides a significant boost in performance. When comparing with 3DVLP with scratch-20, we observe that 3DVLP shows superiority in all metrics with the same training time. The training in the pre-training stage enhances the performance by 0.5-6\% in captioning metrics and 2-4\% in QA metrics.
When comparing with scratch-full, 3DVLP achieves better performance with fewer training times, further verifying the effectiveness of our pre-trained proxy tasks. Interestingly, models mainly develop their captioning ability to high-quality proposals in the late training, as shown by the comparison between scratch-20 and scratch-full.

\section{More Qualitative Results}\label{visual}
We provide more qualitative results in dense captioning in Fig~\ref{cap}.
\section{t-SNE Visualization of Proposal Features}\label{tsne}
We present a t-SNE \cite{van2008visualizing} visualization of proposal features in the scene, as shown in Fig. \ref{tsne_visual}. We use a threshold near the real object center and filter out the proposals representing the background. Furthermore, we assign labels to the proposals with the nearest real object id. We compare the performance of 3DVLP with its variant that does not include OID, OSC, and OCC modules, namely 3DVLP-base. The visualization shows that the object detector in 3DVLP, with the three proposed modules, is better at distinguishing objects in the scene, which facilitates the optimization of downstream tasks.
\end{document}